\pdfoutput=1
\documentclass[11pt]{article}
\usepackage{emnlp2021}

\usepackage{times}
\usepackage{latexsym}

\usepackage{inputenc}
\usepackage{multirow}
\usepackage{times}
\usepackage{enumitem}
\usepackage[running]{lineno}
\usepackage{todonotes}
\usepackage{tcolorbox}
\usepackage{color,soul}
\usepackage[section]{placeins}
\usepackage{epstopdf}
\usepackage{multirow,tabularx}

\title{MIPE: A Metric Independent Pipeline for Effective Code-Mixed NLG Evaluation}

\RequirePackage{colortbl}
\definecolor{airforceblue}{rgb}{0.36, 0.54, 0.66}
\definecolor{amaranth}{rgb}{0.9, 0.17, 0.31}
\definecolor{applegreen}{rgb}{0.55, 0.71, 0.0}
\definecolor{alizarin}{rgb}{0.82, 0.1, 0.26}
\definecolor{azure}{rgb}{0.0, 0.5, 1.0}
\definecolor{cadmiumgreen}{rgb}{0.0, 0.42, 0.24}
\newcommand
\samethanks[1][\value{footnote}]{\footnotemark[#1]}

\author{\hspace{2cm}Ayush Garg\thanks{\hspace{0.2cm}equal contribution}\hspace{0.1cm}, Sammed S Kagi\samethanks\\
    \hspace{2cm}IIT Gandhinagar, India\\
    \hspace{2cm}\small{\{ayush.g,sammed.shantinath\}@iitgn.ac.in}\\\And
    \hspace{2.5cm}Vivek Srivastava\\
    \hspace{2.5cm}TCS Research, India\\
    \hspace{2.5cm}\small{srivastava.vivek2@tcs.com}\\\And
    \hspace{1cm}Mayank Singh\\
    \hspace{1cm}IIT Gandhinagar, India\\
    \hspace{1cm}\small{singh.mayank@iitgn.ac.in}
}

\begin{document}
\maketitle

\begin{abstract}
Code-mixing is a phenomenon of mixing words and phrases from two or more languages in a single utterance of speech and text. Due to the high linguistic diversity, code-mixing presents several challenges in evaluating standard natural language generation (NLG) tasks. Various widely popular metrics perform poorly with the code-mixed NLG tasks. To address this challenge, we present a metric independent evaluation pipeline \textit{MIPE} that significantly improves the correlation between evaluation metrics and human judgments on the generated code-mixed text. As a use case, we demonstrate the performance of \textit{MIPE} on the machine-generated Hinglish (code-mixing of Hindi and English languages) sentences from the \textit{HinGE} corpus. We can extend the proposed evaluation strategy to other code-mixed language pairs, NLG tasks, and evaluation metrics with minimal to no effort. 

\end{abstract}

\section{Introduction}
Code-mixing (hereafter \textit{`CM'}) is a commonly observed communication pattern for a multilingual speaker to mix words and phrases from multiple languages. CM is widespread across various language pairs across the globe, such as Spanish-English (Spanglish) and Hindi-English (Hinglish). Various studies \cite{lang_growth} have predicted the high growth in the number of CM speakers, which would surpass the number of native speakers in various globally popular languages (e.g., English).

With the advent of social-media platforms (e.g., Twitter, Facebook, etc.), we observe a manifold increase in the CM communication by multilingual speakers. This leads to a large scale availability of CM data for various NLP tasks.  Recently, we witness magnitude of work to address various CM NLP tasks such as language identification~\cite{shekhar2020language,singh2018language, ramanarayanan2019automatic, barman2014code, gundapu2018word}, POS tagging~\cite{singh2018twitter, vyas2014pos, pratapa2018word}, named entity recognition ~\cite{singh2018language, priyadharshini2020named, winata2019learning}, word normalisation \cite{singh2018automatic, parikh2021normalization}, CM metrics \cite{guzman2017metrics, srivastava2021challenges}, sentiment analysis~\cite{patwa2020semeval, joshi2016towards}, stance detection \cite{10.1145/3371158.3371226, sane2019stance}, natural language inference \cite{khanuja2020new}, machine translation \cite{srivastava2020phinc, dhar2018enabling}, and question-answering \cite{chandu2019code, thara2020code}.

We observe a growing interest in the computational linguistic community to study the CM NLG tasks. Recently, various resources and systems have been proposed that explore different dimensions of the CM NLG \cite{yang2020csp, gautam2021comet, gupta2021training, rizvi2021gcm, gupta2020semi, jawahar2021exploring}. Evaluation of the CM NLG tasks is challenging due to the high linguistic diversity and lack of standardization. To address this challenge, \newcite{srivastava2021hinge} has proposed \textit{HinGE} corpus for the Hinglish CM text generation and evaluation (see Section \ref{sec: dataset} for details). \textit{HinGE} corpus demonstrates the inefficacy of various widely popular metrics on the CM dataset.

In this paper, we choose five evaluation metrics (see Section \ref{sec: MIPE} for details) as discussed in \cite{srivastava2021hinge} to demonstrate the efficacy of \textit{MIPE}.
Our proposed metric independent pipeline (\textit{MIPE}) augments these metrics and addresses four major linguistic bottlenecks: (i) spelling variations, (ii) language switching, (iii) missing words, and (iv) the limited number of reference sentences associated with the CM NLG systems. The main contributions are:
\begin{itemize}
    \item We identify four major reasons for the poor quality performance of various widely popular evaluation metrics for the code-mixed NLG evaluation.
    \item We propose a metric independent evaluation pipeline \textit{MIPE} that addresses the identified bottlenecks in CM NLG evaluation. Furthermore, we show its efficacy in generating highly correlated metric scores against human scores.
\end{itemize}
 
The rest of the paper is organized as follows. In Section \ref{sec: dataset}, we discuss the dataset for the CM NLG evaluation task. In Section \ref{sec: MIPE}, we present the \textit{MIPE} pipeline addressing the four major bottlenecks for effective CM NLG evaluation. We discuss the results in Section \ref{sec: results}. In Section \ref{sec: challenges}, we discuss the current state and future direction. We conclude the discussion in Section \ref{sec: conclusion}.

\section{Dataset}
\label{sec: dataset}
\begin{figure}[!tbh]
\centering
    \includegraphics[width=1.0\linewidth]{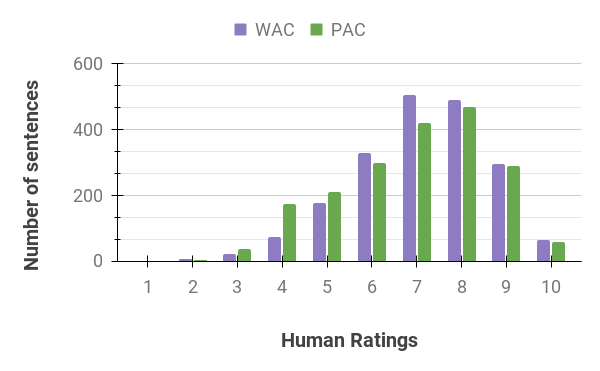}
\caption{Distribution of human ratings on the generated Hinglish sentences using WAC and PAC. The figure is taken from \citet{srivastava2021hinge}.}
\label{fig:ratings}
\end{figure}

Recently, we observe various works to address the underlying challenges with the CM NLG. Numerous resources and systems have been proposed recently to advance the field. In our experiments, we use the \textit{HinGE} corpus proposed in \cite{srivastava2021hinge}. The \textit{HinGE} corpus contains 1,976 English-Hindi parallel sentences from the IIT-B parallel corpus \cite{kunchukuttan2018iit}. Corresponding to each of the English-Hindi parallel sentences, \textit{HinGE} has two variants of CM Hinglish sentences:
\begin{itemize}
    \item Human-generated Hinglish sentences: \cite{srivastava2021hinge} have employed eight human annotators to generate the Hinglish sentences. Each parallel sentence pair is annotated by a single human annotator. Human annotators have generated at least two Hinglish sentences corresponding to each parallel sentence pair. On average, 2.5 Hinglish sentences are generated for each parallel sentence pair.
    \item Machine-generated Hinglish sentences: \newcite{srivastava2021hinge} proposes two rule-based algorithms to generate the CM sentences. They leverage the matrix-frame theory to generate the Hinglish sentences where Hindi is the matrix language and English tokens are embedded. The proposed algorithms differ significantly at the level of granularity (i.e., word and phrase). We will use the acronyms WAC (word-aligned code-mixing) and PAC (phrase-aligned code-mixing) for the two algorithm variants in the rest of the paper.
\end{itemize}
In addition to the machine-generated Hinglish sentences, \textit{HinGE} has a human rating corresponding to each generated sentence. The human rating varies between 1--10, indicating low to high generation quality. Two human annotators have rated each of the machine-generated CM sentences. Figure \ref{fig:example} shows the example CM sentences generated by humans and two rule-based algorithms along with the rating to the machine-generated CM sentences. Figure \ref{fig:ratings} shows the distribution of the human ratings to the machine-generated Hinglish sentences. WAC-generated sentences receive a relatively high rating ($>6$) as compared to PAC. In addition, WAC showed a low degree of human disagreements than PAC.

\begin{figure}[t]
\centering
\small{
\begin{tcolorbox}[colback=white]

\textsc{Human-generated 1}: \textcolor{blue}{is another human being saying, ``kya aapko samaj aaya?''}\\
\textsc{Human-generated 2}: \textcolor{blue}{koi dusra human being yeh kahe, ``Do you understand this?''}\\

\textsc{WAC generated}: \textcolor{orange}{koee doosra human ye kahe, kya aapko samajh aaya} \\
\textsc{Rating 1}: 9\\ 
\textsc{Rating 2}: 8\\

\textsc{PAC generated}: \textcolor{orange}{koee doosra manushy ye kahe, kya aapko samajh understand} \\
\textsc{Rating 1}: 7\\ 
\textsc{Rating 2}: 8
\end{tcolorbox}}
\caption{Example of the CM sentences generated by the annotator along with WAC and PAC generated CM sentence. Two human annotators rate the machine-generated sentence on a scale of 1--10.}
\label{fig:example}
\end{figure}

\noindent\textbf{Efficacy of NLG Evaluation Metrics:}
\newcite{srivastava2021hinge} present a study demonstrating the inefficacy of five widely popular NLG evaluation metric on the \textit{HinGE} corpus. The five metrics are: (i) Bilingual Evaluation Understudy Score (\textbf{BLEU},~\citet{papineni2002bleu}), (ii) \textbf{NIST}~\cite{doddington2002automatic}, (iii) BERTScore (\textbf{BS},~\citet{zhang2019bertscore}), (iv) Word Error Rate (\textbf{WER},~\citet{levenshtein1966binary}), and (v) Translator Error Rate (\textbf{TER},~\citet{snover2006study}). Higher BLEU, NIST, or BS values and lower WER or TER values represent better generation performance.
Tables~\ref{tab: WAC_MIPE} and~\ref{tab: PAC_MIPE} show the comparison of five metric scores against the human ratings against WAC and PAC (see scores present in columns with heading \textit{`Without MIPE'}). In addition, \cite{srivastava2021hinge} present a correlation study between the human ratings and the metric scores. For this purpose they divide the ratings into three buckets:
\begin{itemize}
    \item Bucket 1: Human rating between 2--10.
    \item Bucket 2: Human rating between 2--5.
    \item Bucket 3: Human rating between 6--10.
\end{itemize}

Table \ref{corr} shows the correlation between the human ratings and the metric scores for WAC and PAC (see scores present in columns with heading \textit{`Without MIPE'}). The correlation scores show a scope to build systems that shows a high correlation with human judgment.

\section{MIPE}
\label{sec: MIPE}
\begin{figure*}[!tbh]
\centering
\includegraphics[width=1\linewidth]{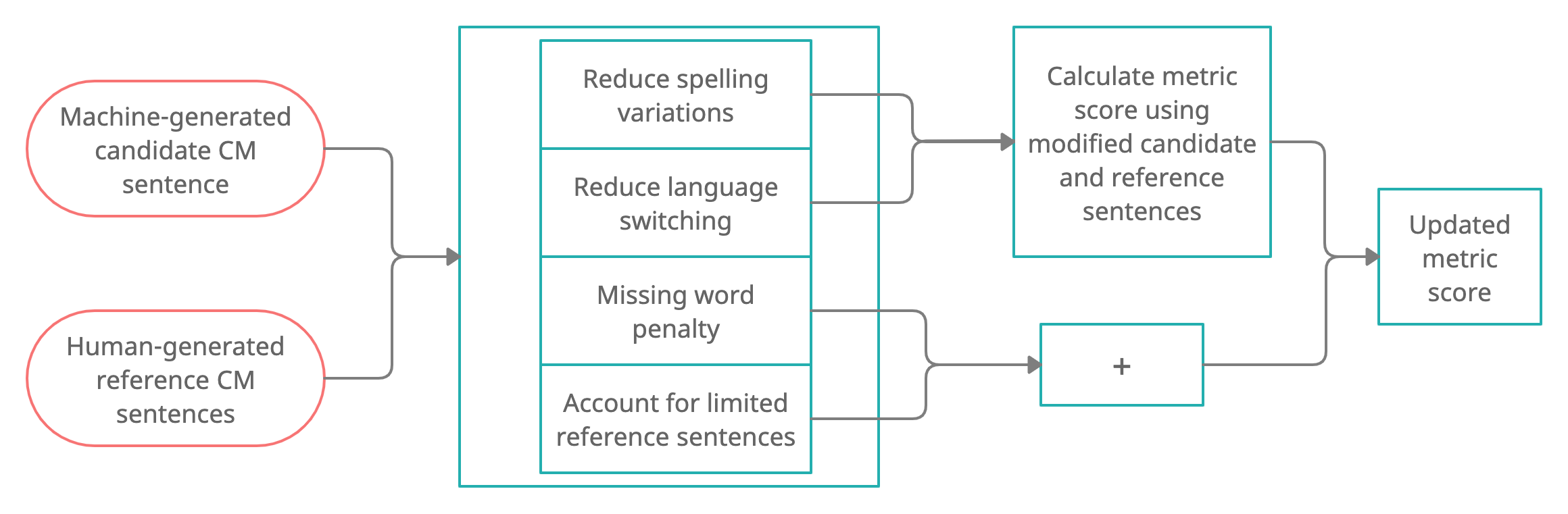}
\caption{The architecture of the proposed CM NLG evaluation system \textit{MIPE}. Machine-generated candidate CM sentences are generated by two rule-based algorithms (WAC and PAC). We reduce the spelling variation and language switching for both the candidate and reference sentences based on phonetics. A penalty is applied to the words in the candidate sentence which are not present in any of the reference sentences. We account for limited reference sentences by chunking the candidate and reference sentences into trigram phrases. The words in the candidate trigram phrases are assigned a score based on their presence in the reference phrases. The candidate phrase score is used to account for the paraphrasing possibilities of the CM reference sentences.}
\label{fig:textmate}
\end{figure*}

As discussed in the previous section, the widely used evaluation metrics fail to capture the linguistic diversity of the CM data. Based on the empirical observation on the 10 datasets used in \cite{srivastava2021challenges}, we identify four major reasons for the failure of NLG evaluation metrics on the CM data. We propose a metric independent evaluation pipeline \textit{MIPE}, for effective evaluation. Using \textit{MIPE}, we first reduce the spelling variations (see Section~\ref{sec: spell}) and the language switching (see Section~\ref{SWS})  in the candidate Hinglish sentence. Next, we introduce a penalty (see Section~\ref{missing words}) on the evaluation score based on the degree of importance of the missing words in the candidate Hinglish sentence. Finally, we address the challenge of a limited number of reference sentences (see Section~\ref{limited}) by segmenting the candidate and the reference sentences into phrases and leveraging the paraphrasing capability. Figure \ref{fig:textmate} shows the architecture of the proposed evaluation pipeline.

\subsection{Spelling variations}
\label{sec: spell}
The first challenge to effective evaluation is the non-standard spellings of the code-mixed words. E.g., words \textit{kanekt}, \textit{connect}, and \textit{connekt} conveys the same meaning in a Hinglish sentence. Due to a lack of writing standards for the code-mixed languages, the speakers often use their phonetic understanding of the source languages to write the CM sentences. Hence, in most spelling variations, the addition, omission, and substitution of letters indicate that the phonetics remains almost the same. Specifically, we observe three major reasons for spelling variations,
\begin{itemize}
    \item $R1$: character repetition
    \item $R2$: replacement with similar-sounding character
    \item $R3$: vowel omission
\end{itemize}
To address these problems, we normalize words such that similar-sounding words are grouped.
We leverage the concept of Phonetic Dissimilarity (PDS,~\citet{toutanova2002pronunciation}) to address the spelling variations in the CM language. Our proposed PDS algorithm is a variant of the popular dynamic programming-based edit distance algorithm. Similar to edit distance, PDS quantifies the dissimilarity between two strings by counting the minimum number of edit operations (addition, deletion, and substitution) required to transform one string into the other. In PDS, we assign different costs to each edit operation based on the phonetic characteristics of the corresponding characters of the two words and the edit operation under consideration. To access the phonetic characteristics, we use a corpus of all possible pronunciations of the English alphabets\footnote{\url{https://www.speakmethod.com/alphabet\_sounds.html}}. Algorithm 1
describes PDS between a word $w_1$ (in candidate CM sentence) and $w_2$ (in reference CM sentences).  To address $R1$, we remove repeating characters from both words. By default, we keep addition and deletion cost = 1 and substitution cost = 2. To address $R2$, we decrease the substitution cost to $\rho_{sub}$ for similar-sounding characters as substitution of one of these characters is highly likely. To address $R3$, we decrease the addition cost of vowels to $\rho_{add}$ and the deletion cost of vowels to $\rho_{del}$, where $\rho_{add} > \rho_{del}$. This is due to the empirical observation that the omission of vowels is much more likely than an addition. Further, we decrease the addition and deletion costs of a possible silent character to $\rho_{sil}$. We consider the minimum of PDS($w1$, $w2$) and PDS($w2$, $w1$) as the final PDS score to identify the spelling variation between words $w1$ and $w2$. In our experiments, we keep  $\rho_{sub} = \rho_{add} = \rho_{sil} = 0.75$, and $\rho_{del} = 0.25$.


     
     


\begin{figure}[!tbh]
\centering
\includegraphics[width=1.0\linewidth]{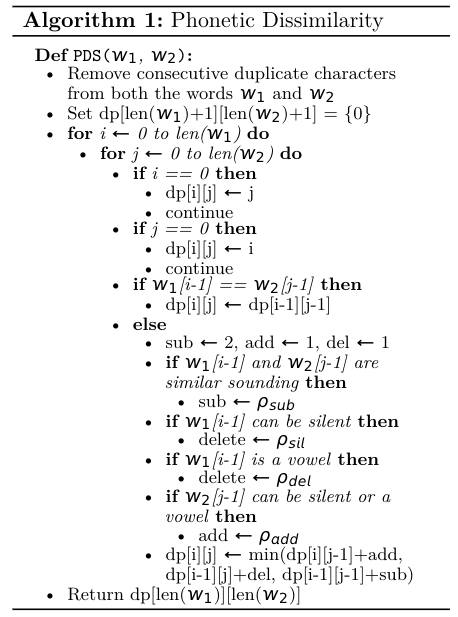}
\label{Phonetic Dissimilarity}
\end{figure}

\subsection{Similar Words}
\label{SWS}
Identifying similar words in the same or different languages is a challenging task in the CM languages. For example, two phrases \textit{``in the market''} and \textit{``in the bazaar''} convey the same semantics, but most automatic evaluation metrics will fail to identify the semantic similarity. To address the challenge of token-level similarity, we need a common representation of words across the source languages.  
To mitigate this problem, we propose a Similar Word Search (SWS) procedure. Algorithm 2 
shows the description of the SWS procedure. Given a word from the candidate CM sentence as an input, the SWS procedure returns all similar words from the corresponding reference sentences. We select that word from the reference sentences, which yields the minimum PDS value. The SWS procedure returns a word from the reference set if the minimum PDS value is less than $\sigma_{thres}$. Otherwise, it computes pairwise cosine distance (in the cross-lingual word embedding space) between each word in a set of reference words and the input word. To create the cross-lingual embedding space, we use the pre-trained word vectors of dimension 300 for English and Hindi from fastText \cite{bojanowski2017enriching}. For the shared representation, we use VecMap \cite{artetxe2018acl} to learn the mapping in an unsupervised fashion with the default settings. We use the English and Hindi sentences from the IIT-B parallel corpus \cite{kunchukuttan2018iit}. In case the cosine similarity is greater than $\sigma_{cos}$, the SWS procedure returns the word from the reference set; else, we assume that no similar word exists in the reference set. In our experiments, we keep  $\sigma_{thres} =2$, and $\sigma_{cos}=0.5$.


    

\begin{figure}[!tbh]
\centering
\includegraphics[width=1.0\linewidth]{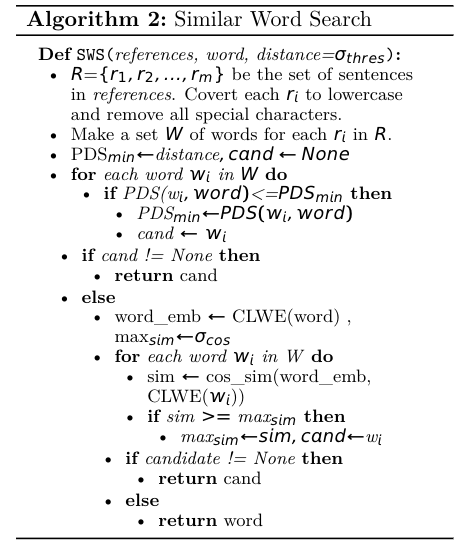}
\label{Similar Word Search}
\end{figure}

\subsection{Missing words}
\label{missing words}
Generally, the generated candidate sentence misses some words resulting in a significant impact on the automatic evaluation scores. Some words are more important than others, but most metrics consider them equal ($M1$). Furthermore, most metrics match exact words with no flexibility in spelling variations and language switching ($M2$). Here, we address both these problems to apply a missing word penalty to the metric score with some writing style flexibility.
To address $M1$, we use WAC procedure\footnote{We employ WAC due to its capability to generate high-quality sentences (as shown in \cite{srivastava2021hinge}). Also, the Hinglish sentence generated by WAC has words from only the source English and Hindi sentences which in turn doesn't influence the IDF values of the generated words to a large extent.} to generate a large Hinglish corpus (hereafter \textit{`ParallelCorp'}) of 2,132,184 sentences. For creating the parallel corpus, we collect English sentences from multiple sources\footnote{\url{https://www.kaggle.com/kazanova/sentiment140}}\textsuperscript{,}\footnote{\url{https://www.kaggle.com/arkhoshghalb/twitter-sentiment-analysis-hatred-speech}}\textsuperscript{,}\footnote{\url{http://www.cfilt.iitb.ac.in/iitb_parallel/}}\textsuperscript{,}\footnote{\url{https://www.kaggle.com/columbine/imdb-dataset-sentiment-analysis-in-csv-format}} and translate them (if not already translated) into Hindi language using Google Translate API. We calculate IDF-values (Inverse Document Frequency) of each word in the Hinglish corpus. The words with low IDF values occur rarely and hence carry more semantic information. If a word is not present in the \textit{ParallelCorp}, we consider it semantically important. To address $M2$, we relax the exact match condition by postulating that either the word is present in the candidate sentence or its variant is present in the sentence. Here, we allow two types of variations (i) minor spelling variations and (ii) language switch (for more details, see Sections~\ref{sec: spell} and~\ref{SWS}). We use the SWS procedure to find a word variant keeping a maximum distance value of 1. Algorithm 3 
shows the description of the missing word penalty (MWP) in detail. For each word $w$ in a reference sentence, we check the presence of $w$ and its variants in the candidate sentence. In case $w$ is not present, we add $w$'s IDF value as the penalty for the absence. We repeat the procedure for each reference sentence and take the minimum penalty among all reference sentences. We reduce the MWP score from the metric score for a given evaluation metric.



\begin{figure}[!tbh]
\centering
\includegraphics[width=1.0\linewidth]{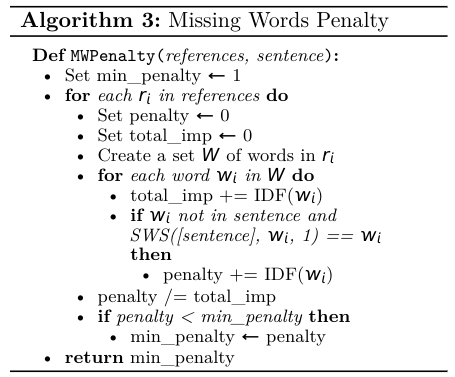}
\label{algo: MWP}
\end{figure}

\subsection{Limited Reference Sentences}  
\label{limited}
A sentence can be paraphrased in numerous ways by interchanging subject and predicate, active and passive voice, and first, second, and third-person perspectives. With the code-mixed text, the paraphrasing possibilities significantly increase. For an automatic evaluation, it is infeasible to generate all possible paraphrases as reference sentences. Even though \textit{HinGE} dataset has at least two reference sentences against a candidate sentence but it is insufficient to include all the possibilities. Thus, paraphrasing drastically limits the evaluation capabilities of various metrics. To address this problem, we present an algorithm \textit{PhraseScore} that captures the paraphrasing capability of the reference sentences. Algorithm 4 
shows the description of the \textit{PhraseScore} method. We split the candidate sentence and the set of reference sentences into trigram phrases. If word \textit{w} in the candidate phrase exists in one of the reference phrases, we add the IDF value of the \textit{w} in the phrase score for that phrase. Else, we subtract the IDF value as a penalty. This phrase score is aggregated, normalized over the number of phrases in the candidate sentence, and divided by the penalty of missing words in the candidate sentence. To prevent division by zero, we add 0.0001 to the penalty. In case a word is not present in the IDF dictionary, we assign it a relatively high value ($\mu_{miss}$) to indicate that it is a rare word of high importance. Finally, we increase the metric score by adding the candidate sentence's \textit{PhraseScore}.  In our experiments, we keep $\mu_{miss}$=20. Due to the unavailability of a paraphrasing system for a code-mixed language, the formulation of \textit{PhraseScore} algorithm depends on the assumption that the trigram phrases in a sentence can be reordered to create new sentences.



\section{Results and Evaluation}
\label{sec: results}
We evaluate WAC and PAC procedures augmented with \textit{MIPE} pipeline against all the five metrics (as discussed in Section \ref{sec: dataset}). Table \ref{tab: WAC_MIPE} and \ref{tab: PAC_MIPE} shows the effect of \textit{MIPE} against the five metrics. As expected, all metrics show better scores with the \textit{MIPE} augmentation. The metric scores after the \textit{MIPE} shows a high correlation\footnote{We experiment with Pearson Correlation Coefficient.} against the metric scores without \textit{MIPE} (see Table \ref{tab: corr_with}). This shows that improvements in the metric scores is constant throughout and are not by chance. 

\begin{figure}[!tbh]
\centering
\includegraphics[width=1.0\linewidth]{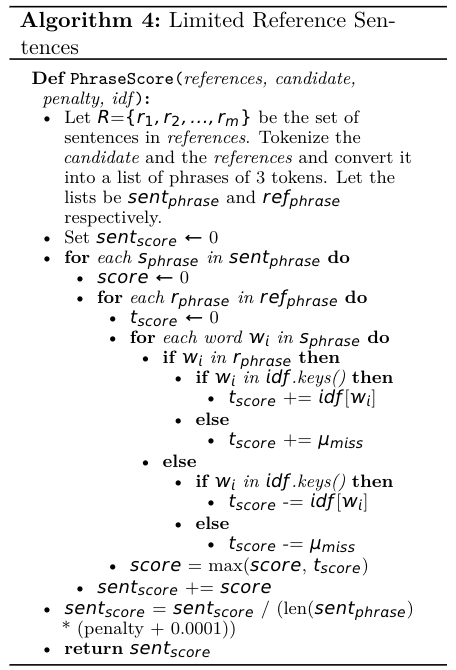}
\label{algo:PhraseScore}
\end{figure}

Table~\ref{corr} shows the effect of \textit{MIPE} on the correlation with the human scores. We use the same criteria to bucket the human ratings as discussed in Section \ref{sec: dataset}. We observe a higher correlation in all the three buckets for WAC augmented with \textit{MIPE}. This improvement is consistent throughout all the metrics. For PAC augmented with \textit{MIPE}, we observe a decrease in correlation in the second bucket, which can be attributed to (i) a relatively large number of poor quality (low human scores) sentences generated by PAC, and (ii) rating poor quality CM sentence is a challenging task for humans due to lower readability of the sentence.
For the rest of the buckets, PAC with \textit{MIPE} shows a higher correlation with the human scores.

\section{Current State and Future Directions}
\label{sec: challenges}

\begin{table*}[!tbh]
\centering
\small{
\resizebox{\hsize}{!}{
\begin{tabular}{|c|c|c|c|c|c|c|c|c|c|c|} 
\cline{1-11}
\multirow{2}{*}{\begin{tabular}[c]{@{}c@{}}\textbf{Human }\\\textbf{ score}\end{tabular}} & \multicolumn{5}{c|}{\textbf{Without \textit{MIPE}}}                       & \multicolumn{5}{c|}{\textbf{With \textit{MIPE}}}                          \\ 
\cline{2-11}                      & \textbf{BLEU} & \textbf{WER} & \textbf{TER} & \textbf{NIST} & \textbf{BS} & \textbf{BLEU} & \textbf{WER} & \textbf{TER} & \textbf{NIST} & \textbf{BS}  \\ 
\hline
\textbf{2}                                                                                & 0.144         & 0.741        & 0.667        & 0.092         & 0.851       & 0.238         & 0.651        & 0.544        & 0.140         & 0.860        \\
\textbf{3}                                                                                & 0.138         & 0.735        & 0.708        & 0.070         & 0.852       & 0.323         & 0.625        & 0.569        & 0.133         & 0.860        \\
\textbf{4}                                                                                & 0.133         & 0.695        & 0.666        & 0.103         & 0.849       & 0.391         & 0.536        & 0.480        & 0.184         & 0.906        \\
\textbf{5}                                                                                & 0.135         & 0.711        & 0.681        & 0.110         & 0.853       & 0.380         & 0.556        & 0.494        & 0.172         & 0.985        \\
\textbf{6}                                                                                & 0.141         & 0.697        & 0.670        & 0.102         & 0.852       & 0.361         & 0.560        & 0.502        & 0.144         & 0.967        \\
\textbf{7}                                                                                & 0.161         & 0.663        & 0.630        & 0.111         & 0.856       & 0.398         & 0.522        & 0.453        & 0.168         & 0.947        \\
\textbf{8}                                                                                & 0.177         & 0.621        & 0.589        & 0.127         & 0.859       & 0.465         & 0.445        & 0.377        & 0.204         & 0.976        \\
\textbf{9}                                                                                & 0.212         & 0.571        & 0.538        & 0.150         & 0.865       & 0.531         & 0.387        & 0.313        & 0.242         & 1.000        \\
\textbf{10}                                                                               & 0.291         & 0.509        & 0.493        & 0.157         & 0.878       & 0.572         & 0.318        & 0.252        & 0.291         & 1.000        \\
\hline
\end{tabular}}}
\caption{Comparison of metric scores with and without using \textit{MIPE} for WAC.}
\label{tab: WAC_MIPE}
\end{table*}

\begin{table*}[!tbh]
\centering
\small{
\resizebox{\hsize}{!}{
\begin{tabular}{|c|c|c|c|c|c|c|c|c|c|c|} 
\cline{1-11}
\multirow{2}{*}{\begin{tabular}[c]{@{}c@{}}\textbf{Human }\\\textbf{ score}\end{tabular}} & \multicolumn{5}{c|}{\textbf{Without \textit{MIPE}}}                       & \multicolumn{5}{c|}{\textbf{With \textit{MIPE}}}                           \\ 
\cline{2-11}                    & \textbf{BLEU} & \textbf{WER} & \textbf{TER} & \textbf{NIST} & \textbf{BS} & \textbf{BLEU} & \textbf{WER} & \textbf{TER} & \textbf{NIST} & \textbf{BS}  \\ 
\hline
\textbf{2}                                                                                & 0.126         & 0.672        & 0.698        & 0.176         & 0.8603      & 0.338         & 0.474        & 0.500        & 0.318         & 0.997        \\
\textbf{3}                                                                                & 0.146         & 0.765        & 0.696        & 0.086         & 0.851       & 0.425         & 0.603        & 0.526        & 0.120         & 0.883        \\
\textbf{4}                                                                                & 0.143         & 0.744        & 0.703        & 0.100         & 0.8464      & 0.419         & 0.598        & 0.523        & 0.130         & 0.888        \\
\textbf{5}                                                                                & 0.153         & 0.726        & 0.680        & 0.114         & 0.8515      & 0.407         & 0.589        & 0.508        & 0.154         & 0.894        \\
\textbf{6}                                                                                & 0.164         & 0.689        & 0.646        & 0.124         & 0.8558      & 0.449         & 0.525        & 0.456        & 0.171         & 0.912        \\
\textbf{7}                                                                                & 0.176         & 0.661        & 0.618        & 0.121         & 0.8581      & 0.475         & 0.485        & 0.411        & 0.198         & 0.936        \\
\textbf{8}                                                                                & 0.177         & 0.639        & 0.605        & 0.128         & 0.8598      & 0.498         & 0.437        & 0.370        & 0.200         & 0.938        \\
\textbf{9}                                                                                & 0.184         & 0.614        & 0.590        & 0.129         & 0.8638      & 0.545         & 0.387        & 0.321        & 0.230         & 0.967        \\
\textbf{10}                                                                               & 0.242         & 0.551        & 0.543        & 0.146         & 0.8731      & 0.600         & 0.314        & 0.262        & 0.280         & 0.997        \\
\hline
\end{tabular}}}
\caption{Comparison of metric scores with and without using \textit{MIPE} for PAC.}
\label{tab: PAC_MIPE}
\end{table*}

\begin{table*}[!tbh]
\centering
\small{
\begin{tabular}{|c|c|c|c|c|c|}
\hline
   & \textbf{BLEU} & \textbf{WER} & \textbf{TER} & \textbf{NIST} & \textbf{BS} \\ \hline
\textbf{WAC} & 0.948    & 0.988   & 0.984   & 0.961 & 0.8326 \\ \hline
\textbf{PAC} & 0.830    & 0.986   & 0.982   & 0.944 & 0.8843 \\ \hline
\end{tabular}}
\caption{Correlation between the evaluation metric scores with and without using \textit{MIPE} pipeline.}
\label{tab: corr_with}
\end{table*}

\begin{table*}[!tbh]
\small{
\resizebox{\hsize}{!}{
\begin{tabular}{|c|c|c|c|c|c|c|c|c|c|c|c|c|}
\hline
\multirow{3}{*}{} & \multicolumn{6}{c|}{\textbf{\begin{tabular}[c]{@{}c@{}}Correlation with human scores \\ (Without MIPE)\end{tabular}}}  & \multicolumn{6}{c|}{\textbf{\begin{tabular}[c]{@{}c@{}}Correlation with human scores \\ (With MIPE)\end{tabular}}}    \\ \cline{2-13} 
   & \multicolumn{2}{c|}{\textbf{Bucket 1}}  & \multicolumn{2}{c|}{\textbf{Bucket 2}} & \multicolumn{2}{c|}{\textbf{Bucket 3}} & \multicolumn{2}{c|}{\textbf{Bucket 1}} & \multicolumn{2}{c|}{\textbf{Bucket 2}} & \multicolumn{2}{c|}{\textbf{Bucket 3}} \\ \cline{2-13} 
   & \textbf{WAC} & \textbf{PAC}  & \textbf{WAC}    & \textbf{PAC}  & \textbf{WAC} & \textbf{PAC} & \textbf{WAC}  & \textbf{PAC}     & \textbf{WAC}  & \textbf{PAC}    & \textbf{WAC} & \textbf{PAC} \\ \hline
\textbf{BLEU}     & 0.810   & 0.910    & -0.861     & \textbf{0.878}     & 0.941   & 0.844   & \textbf{0.942}     & \textbf{0.950}   & \textbf{0.910}     & 0.643      & \textbf{0.994}    & \textbf{0.981}    \\ \hline
\textbf{WER} & -0.936  & \textbf{-0.822}    & -0.785     & \textbf{0.457}     & -0.993  & -0.973  & \textbf{-0.949}    & -0.780      & \textbf{-0.880}    & 0.713      & \textbf{-0.995}   & \textbf{-0.993}   \\ \hline
\textbf{TER} & -0.891  & \textbf{-0.963}    & 0.000      & \textbf{-0.610}    & -0.998  & -0.970  & \textbf{-0.932}    & -0.937      & \textbf{-0.737}    & 0.229      & \textbf{-0.998}   & \textbf{-0.997}   \\ \hline
\textbf{NIST}     & \textbf{0.913}    & 0.127    & 0.642      & \textbf{-0.559}    & 0.986   & 0.846   & 0.851    & \textbf{0.246}   & \textbf{0.769}     & -0.671     & \textbf{0.993}    & \textbf{0.952}    \\ \hline
\textbf{BS} &  0.844  &  \textbf{0.710}   &  0.227   &   \textbf{-0.689}    &   \textbf{0.953}  &  0.937  &   \textbf{0.924}   & 0.400   &   \textbf{0.922}   & -0.720     &   0.895  & \textbf{0.972}    \\ \hline
\end{tabular}}}
\caption{Comparison of correlation between evaluation metrics and human scores for WAC and PAC with and without \textit{MIPE} pipeline.}
\label{corr}
\end{table*}

The results discussed in Section \ref{sec: results} demonstrate a need to build metrics, theories, and experiments for better CM NLG evaluation. Some of the challenges and limitations of the proposed \textit{MIPE} pipeline for effective CM NLG includes:
\begin{itemize}
    \item Due to the unavailability of resources in other CM language pairs, the \textit{MIPE} pipeline is tested on a single CM language. We need to extend the proposed evaluation strategy to other CM language pairs.
    \item The presence of two different languages in a single CM sentence increases the paraphrasing possibility to a much larger extent. We need metrics that attend to the CM sentences beyond the bag of words model. These metrics should also be able to account for paraphrasing.
    \item There are various other reasons (beyond the four reasons discussed in this paper) that influence the evaluation of CM NLG tasks such as named-entities, transliteration, etc. The \textit{MIPE} pipeline doesn't currently account for these limitations.
    \item The code-mixed sentences in the \textit{HinGE} dataset are not collected from the social media platforms. The code-mixed data from the social media platform tends to be more noisy and distorted which could influence the performance of \textit{MIPE} pipeline.
\end{itemize}

As discussed, currently there are several limitations with the CM NLG evaluation which need to be addressed in order to build effective CM NLG systems for multilingual societies. Some of the lessons learned and the future directions for the CM NLG evaluations are:

\begin{itemize}
    \item The limited resource availability is one of the major bottlenecks in the CM NLG tasks and evaluation. Currently, the available resources are smaller in size compared to the monolingual NLG tasks.
    \item In contrast to the \textit{MIPE} augmentation pipeline, we need systems that can leverage the noisy nature of the code-mixed text. The currently proposed \textit{MIPE} pipeline addresses the various challenges independently and attempts to reconstruct the noisy CM text for effective evaluation.  
    \item The two languages participating in CM influence the various constructs of the target CM sentence such as grammar, syntax, etc. The current experimentation with only one CM language needs to be explored with other CM languages.  
    \item Recently, we observe a rise in the availability of multilingual language models (LMs). These LMs could be used to build effective CM NLG evaluation systems.
    \item The current evaluation metrics seem to perform poorly with the CM languages. We need to build dedicated metrics for the CM NLG evaluation tasks that can leverage the linguistic diversity of the CM data.
\end{itemize}

\section{Conclusion}
\label{sec: conclusion}
In this paper, we present a metric independent evaluation pipeline for efficient code-mixed NLG evaluation. The proposed pipeline shows a high correlation between the human scores and the underlying evaluation metrics. Besides the four significant challenges to CM NLG evaluation, in the future, we also plan to address other challenges such as code-mixed existence of named-entities, informal writing style, and missing context. 

\bibliography{anthology,custom}
\bibliographystyle{acl_natbib}

\end{document}